\pdfoutput=1

\documentclass[11pt]{article}

\usepackage{EMNLP2023}

\usepackage{times}
\usepackage{latexsym}

\usepackage[T1]{fontenc}

\usepackage[utf8]{inputenc}

\usepackage{microtype}

\usepackage{inconsolata}

\usepackage{makecell}
\usepackage{multirow}
\usepackage{booktabs}
\usepackage{amsmath}
\usepackage{amsfonts}
\usepackage{mathtools}
\usepackage{caption}
\usepackage{subcaption}

\usepackage{graphicx}
\graphicspath{{figs/}}
\usepackage{xcolor}
\usepackage{placeins}

\newcommand{\greenemph}[1]{\textcolor{green!70!black}{\textbf{#1}}}
\newcommand{\greenemphmath}[1]{{\color{green!70!black}{#1}}}
\newcommand{\redemph}[1]{\textcolor{red!70!black}{\textbf{#1}}}
\newcommand{\redemphmath}[1]{{\color{red!70!black}{#1}}}
\newcommand{\greenmark}[1]{\colorbox{green!30}{#1}}
\newcommand{\redmark}[1]{\colorbox{red!30}{#1}}

\newcommand{\quash}[1]{}  
\newcommand{\cready}[1]{}  

\title{MuLER: Detailed and Scalable Reference-based Evaluation}

\author{Taelin Karidi\\  Hebrew University of Jerusalem \\ {\small \href{mailto:taelin.karidi@mail.huji.ac.il}{\tt taelin.karidi@mail.huji.ac.il}}  
\And Leshem Choshen \\ Hebrew University of Jerusalem \\ {\small \href{mailto:leshem.choshen@mail.huji.ac.il}{\tt leshem.choshen@mail.huji.ac.il}} \\  
\AND Gal Patel \\ Hebrew University of Jerusalem \\ {\small \href{mailto:gal.patel@mail.huji.ac.il}{\tt gal.patel@mail.huji.ac.il}} \\  
\And Omri Abend \\ Hebrew University of Jerusalem \\ {\small \href{mailto:omri.abend@cs.huji.ac.il}{\tt omri.abend@cs.huji.ac.il}}
}

\begin{document}

\maketitle

\begin{abstract}

We propose a novel methodology (namely, \textbf{MuLER}) that transforms any reference-based evaluation metric for text generation, such as machine translation (MT) into a fine-grained analysis tool. Given a system and a metric, MuLER quantifies how much the chosen metric penalizes specific error types (e.g., errors in translating names of locations). MuLER thus enables a detailed error analysis which can lead to  targeted improvement efforts for specific phenomena. We perform experiments in both synthetic and naturalistic settings to support MuLER's validity and showcase its usability in MT evaluation, and other tasks, such as summarization.  Analyzing all submissions to WMT in 2014$-$2020, we find consistent trends. For example, nouns and verbs are among the most frequent POS tags. However, they are among the hardest to translate.
Performance on most POS tags improves with overall system performance, but a few are not thus correlated (their identity changes from language to language). 
Preliminary experiments with summarization reveal similar trends.\footnote{Our codebase is found here: https://github.com/tai314159/MuLER}
\end{abstract}
\begin{figure}
\centering
\includegraphics[width=\columnwidth]{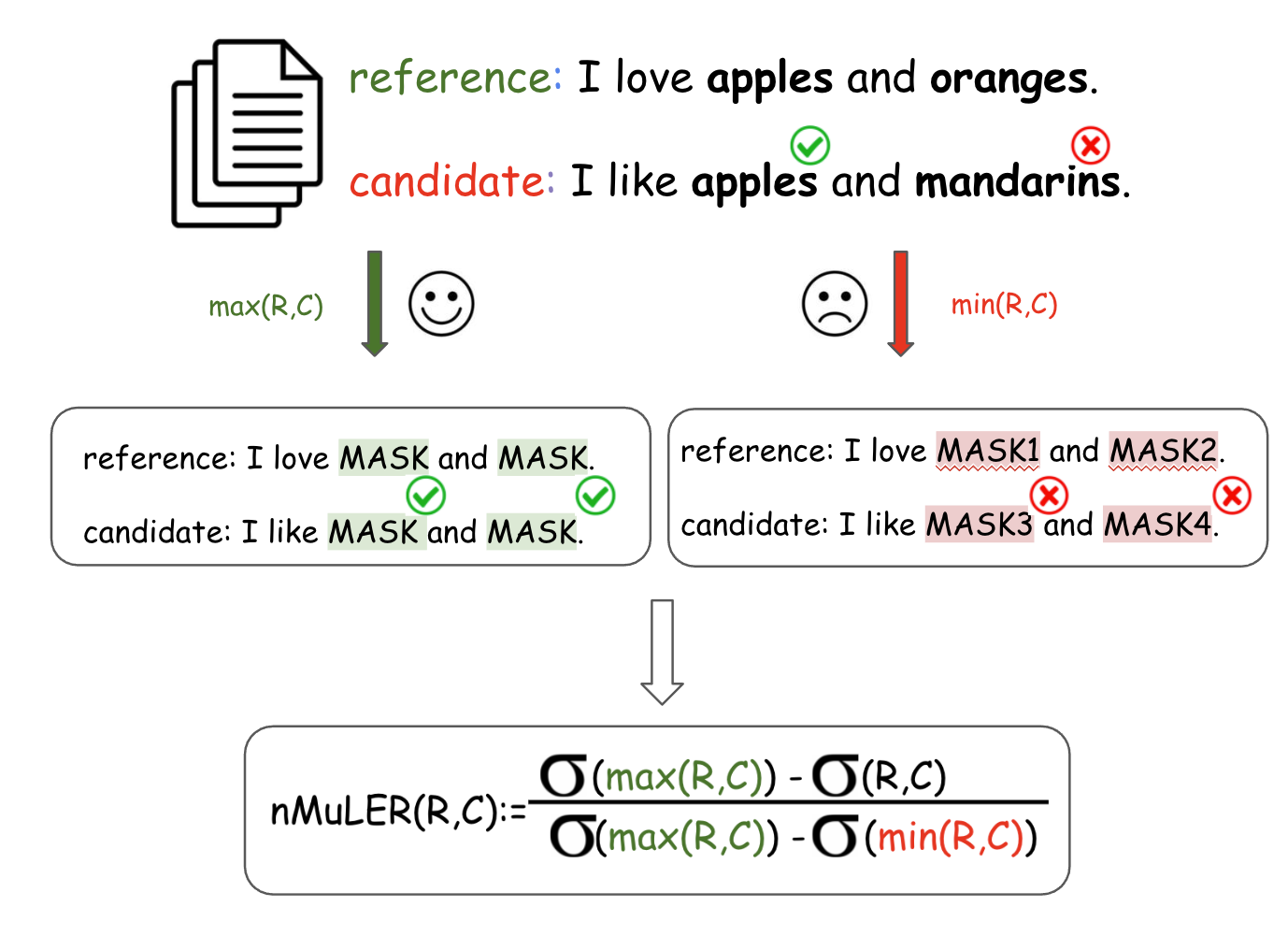}
\caption{Illustration of MuLER for the feature NOUN. Two masking strategies are employed on the reference and the candidate -- Oracle masking $max(R,C)$, and anti-oracle masking $min(R,C)$. $\sigma$ is the task's metric (e.g. BLEU, ROUGE).}  \label{fig:main_fig}
\end{figure}

\section{Introduction}

Reference-based evaluation of text generation plays a uniquely important role in the development of machine translation \citep{papineni-etal-2002-bleu}, summarization \citep{lin2004rouge}, and simplification \citep{sari} among many other sub-fields of NLP. It allows a scalable, cheap evaluation that often correlates at the system-level with human evaluation.

However, reference-based evaluation metrics tend to produce a bottom line score, allowing little to no ability for a fine-grained analysis of the systems' strengths and weaknesses. Such an analysis is important, for example, for targeted development efforts that focus on improving specific phenomena, or for better identifying  scenarios in which the system is reliable \cite{liu-etal-2021-explainaboard}.
We propose a novel evaluation methodology, \textbf{Multi-Level Evaluation with Reference} (MuLER), that presents a detailed picture of text generation system's performance. Our methodology allows to slice the data according to different criteria, such as syntactic or semantic ones. 
Given a feature that can be detected automatically on the target side, and a reference-based metric, MuLER allows to scalably measure the system's performance on words and spans that contain this feature.

MuLER thus yields a decomposition of any evaluation metric, to more focused measurements of the system's performance on span-level and 
word-level features, such as POS tags, named entity types, sentence sentiment etc.
Moreover, the methodology and code can be expanded to features of choice.

In providing a per-phenomenon picture of system performance, MuLER is similar to challenge set approaches to evaluation (see \S\ref{sec:background}). 
However, MuLER takes a more naturalistic approach, and narrows the evaluation to the test examples that contain a particular feature. 

Given an evaluation metric (e.g., BLEU) for a text generation task (e.g., MT) and a feature of interest of the system's output (e.g., performance on adjectives), MuLER operates as follows (see \S\ref{subsec:tagger}): It masks the feature in both the reference and the prediction by the same token (e.g., replace each adjective with a placeholder ``ADJ''). This can be seen as an oracle adaptation to the output, that changes the span with the feature to agree with the reference.
MuLER's score is the (normalized) difference between the metric score over the masked texts and the score over the original ones.

We present results from MT as well as summarization and synthetic paraphrasing. In addition, we perform synthetic experiments to validate MuLER's effectiveness and usability.
Our experiments show that MuLER can measure performance on a particular feature (\S\ref{sec:validation}), and reveal some previously unreported patterns in established MT systems (\S\ref{sec:analysis}). For example, while translation of nouns and verbs improved over the years, translation of named entities improve only for some categories \S\ref{subsec:comparing_models}.\looseness=-1


\section{Methodology} \label{sec:methodology}

The \textbf{MuLER} methodology seeks to gain insight as to the performance of a text generation system $s$ according to a given metric $\sigma$ on instances with the feature $f$. The feature is a dimension along which the system is evaluated  that can be automatically detected given text. Examples of features here may be POS tags, named entity types, morphological categories, among others.

MuLER operationalizes this notion as improvement in the score of $s$ according to $\sigma$, if $s$ would have correctly predicted all instances of this feature. For scale, this improvement is compared to the overall possible improvement (the score is defined in  \S\ref{subsec:muler_score}). To assess that, MuLER creates an oracle where the feature $f$ is perfect and an anti-oracle where it is fully wrong (cf.~\S\ref{subsec:oracle_ao_masking}).

\subsection{Feature Tagger: Formal Definition} \label{subsec:tagger}

Let $f$ be a feature of interest. 
Let $S = \{s_1,...,s_n\}$ be a corpus of output sentences (produced by the evaluated system), $R=\{r_1,...,r_n\}$ be a set of corresponding references, and $C=\{c_1,...,c_n\}$ be a set of corresponding candidates. Let $\tau$ be a function from sentences $x \in S \cup R$ that replaces each span containing a feature $f$ with a special mask token $M_f$ (we assume the spans with the $f$ feature are non-overlapping). Denote the $i$-th token in $\tau(x)$ with $\tau(x)^{(i)}$. Then, for each token $\tau(x)^{(i)}$:
\begin{equation}
    \tau(x)^{(i)} = 
    \begin{cases}
        M_f & 
        \begin{array}{l}\text{if $x^{(i)}$ is part of a span} \\ 
        \text{with the feature $f$} \end{array} \\
        x^{(i)} & \text{otherwise} \\
    \end{cases}\label{eq:tagger_general}
\end{equation}

\subsection{Oracle and Anti-Oracle Masking} \label{subsec:oracle_ao_masking}

Let $\sigma$ be a reference-based evaluation metric that takes sets of system outputs $S$ and references and $R$ and returns a real value. We can define two masking strategies that represent the best possible performance on sub-spans marked by $f$, or the worst performance, by applying $\tau$ to $S$ and $R$.

We refer to the optimistic masking strategy as \textbf{oracle} masking and denote it by 
$$\tau_{\textnormal{max}}(s_1,s_2) = (\tau(s_1),\tau (s_2)).$$ 

This strategy coincides with eq.~\ref{eq:tagger_general}. For example, if we take $f$ to be common nouns:

\vspace{.2cm}    
\fbox{\begin{minipage}{.42\textwidth}
\footnotesize{
\textbf{Reference:} John likes apples and oranges. \\
\textbf{Output:} John loves bananas and apples. \\
\rule{\textwidth}{.4pt}\\
$\tau_{\textnormal{max}}$(\textnormal{reference}) = \textnormal{John likes \greenemph{NOUN} and \greenemph{NOUN}.}\\
$\tau_{\textnormal{max}}$(\textnormal{output}) = \textnormal{John loves \greenemph{NOUN} and \greenemph{NOUN}.}
}
\end{minipage}}

To minimize rather than maximize $\sigma(R,C)$ by masking spans with the feature $f$, we apply different masks to the outputs and the references. This strategy generally decreases $\sigma$, as it deletes existing correspondences between the reference and the outputs. We refer to this masking strategy as \textbf{anti-oracle} masking and denote it with $\tau_{\textnormal{min}}$. 

Repeating the example above (NOUN and NOUN' are different tokens):

\vspace{.2cm}    
\fbox{\begin{minipage}{.42\textwidth}
\footnotesize{
\textbf{reference:} John likes apples and oranges. \\
\textbf{output:} John loves bananas and apples. \\
\rule{\textwidth}{.4pt}\\
$\tau_{\textnormal{min}}$(\textnormal{reference}) = \textnormal{John likes \greenemph{NOUN} and \greenemph{NOUN}.}\\
$\tau_{\textnormal{min}}$(\textnormal{output}) = \textnormal{John loves \redemph{NOUN'} and \redemph{NOUN'}.}
}
\end{minipage}}
    
\vspace{.2cm} Let $I \subseteq \{1,...,n\}$ be the indices for which both $r_i \in R$ and $c_i \in C$ contain a span with the feature $f$. The average score with each oracle would be:
\begin{align*}
{\greenemphmath{\textnormal{max}_{\sigma}(R,C)}} \coloneqq \frac{1}{|I|} {\sum}_{i\in I} \sigma(\tau_{max}(r_i,c_i), \\
 {\redemphmath{\textnormal{min}_{\sigma}(R,C)}} \coloneqq \frac{1}{|I|} {\sum}_{i\in I} \sigma(\tau_{min} (r_i,c_i). 
\end{align*}

\subsection{MuLER Score} \label{subsec:muler_score}

Using these definitions, we may now define the MuLER score. 
We define the \textbf{MuLER score} as:  

\begin{equation}
\label{eq:score_muler_model_analysis}
\begin{split}
& MuLER(R,C) \coloneqq \\ & \frac{\greenemphmath{\textnormal{max}_{\sigma}(R,C)} - \sigma(R,C)}{\greenemphmath{\textnormal{max}_{\sigma}(R,C)} - \redemphmath{\textnormal{min}_{\sigma}(R,C)}}
\end{split}
\end{equation}

We compute MuLER variants only on indices in which both the reference and the output contain $f$ (prevents division by zero). 
Note that lower MuLER score indicates better performance. 

Intuitively, MuLER captures the potential gains obtained by the best $f$, where the numerator of the score captures the absolute gains from improving $f$. MuLER is therefore a unitless metric, that measures how much of the potential gain is realized by improving the generated spans with the feature $f$.

For simplicity of notation, we assume a single reference per sentence, but the formulation generalizes straightforwardly to multi-reference settings. 

\subsection{Normalization Term: Discussion} \label{subsec:beyond_rmuler} 

In this section we provide the motivation behind the normalization term in our score (eq.~\ref{eq:score_muler_model_analysis}).  
MuLER seeks to assess a system's ability per feature exhibited in the text. Ideally, features could be analyzed both in a single system (\S\ref{subsec:model_analysis}) and across systems (\S\ref{subsec:comparing_models}). 
However, the latter may require special treatment. To illustrate this claim, imagine two MT systems, one nearly perfect and another that produces random outputs. The perfect system has little to gain by masking spans of a feature $f$.and hence the numerator of MuLER will be around zero. However, this is also the case for the random system, since there is hardly any margin for improvement. Even if some words are correctly predicted, the malformed context means a low sentence score. 
This hints that the numerator is not comparable between systems with substantially different performance and therefore should be normalized.

In order to better capture the systems' overall performance, we leverage the anti-oracle masking, 
noting that $\sigma(R,C)$ is in the interval $[\textnormal{min}_{\sigma}(R,C),\textnormal{max}_{\sigma}(R,C)]$ (except for edge cases, App.~\S\ref{app:tab:neg_score}). 
The length of this max-min interval can be interpreted as the quality in which the system manages to translate the contexts of spans bearing the feature $f$ (the farther the oracle and the anti-oracle are apart, the better the system is in translating the contexts).
To illustrate this point, consider the two extremes. For a high performing system the distance between $\textnormal{min}_{\sigma}(R,C)$ and $\textnormal{max}_{\sigma}(R,C)$ is expected to be substantial. There is a lot to lose from an error. However, a horrible system will have a small distance as the minimum and the maximum will both be around zero.

\subsection{Leveraging Sentence Scorers} \label{subsec:score}

Often, instead of a tagger, a continuous scoring function is available for $f$. A scorer operates on tokens or sentences to capture a certain aspect of the text (such as sentiment or concreteness). We propose a way to utilize scorers to analyze the system's generation abilities along various dimensions.

Let $\sigma: S \to \mathbb{R}$ be a scoring function,  where $S = \{s_1,...,s_n \}$ is a set of sentences. 
For a set of references $R = \{r_1,...,r_k \}$ and a set of candidates $C = \{c_1,...,c_k\}$, where $c_i$ is the candidate of $r_i$ we define a score $s_{\sigma}$ the following way: 
$$ s_{\sigma}(R,C)\coloneqq \frac{1}{k} {\sum}^{k}_{i=1}( \sigma(r_i) - \sigma(c_i)).$$

\paragraph{Complementing scores.}\label{sec:hallucination} 

MuLER is defined only for sentences in which the reference and the candidate contain the feature $f$. Hence, it checks the quality of generation but not cases of over/under generation. To account for such cases and ensure the system even generates the feature, we define a \textbf{discrepancy breakdown}:

$$\eta(f) = [\eta_1(f),\eta_2(f),\eta_3(f)]$$ The discrepancy breakdown consists of $3$ numbers; \textbf{add} ($\eta_1(f)$), \textbf{hit} ($\eta_2(f)$), and \textbf{miss} ($\eta_3(f)$) scores. $\eta_1(f)$ is the number of sentences in which the feature $f$ appears in the reference more times than it appears the output, $\eta_2(f)$ is the number of sentences in which the feature $f$ appears in the output more times than in the reference and $\eta_3(f)$ is the number of other sentences with equal amount of times.
See \S\ref{subsec:paraphrases_gender} for usage example of the score.

\section{Experimental Setup} \label{sec:experimental_setup}

\paragraph{Evaluation Metrics.} \label{subsec:eval_metrics}
As reference-based metrics, we consider \href{https://www.nltk.org/_modules/nltk/translate/bleu_score.html}{BLEU} \citep{papineni-etal-2002-bleu}, \href{https://github.com/Tiiiger/bert_score}{BERTScore} \citep{zhang2019bertscore} and \href{https://pypi.org/project/rouge-score/}{ROUGE} \cite{lin2004rouge}. BLEU was developed to measure machine translation quality, and focuses on precision. ROUGE is made for summarization and focuses on recall. Both are based on overlapping n-grams, while BERTScore, a metric for text generation quality, is based on similarity between contextualised embeddings. 
For these metrics, the basic unit of evaluation is a sentence, as it compares between a reference sentence (a human translation) and a candidate sentence (an output of a system).

\paragraph{Features.}
\label{subsec:features}

We experiment with several feature types, each separated into different features: POS tagging, NER and dependency features (see App.~\S\ref{app:tab:feat} -- for full description).

\paragraph{Sentence Scorers.}
As dedicated scorers, we look at sentiment analysis, concreteness, valence, dominance and arousal (cf. App.~
\ref{ap:sec:scorers}.)  

\paragraph{Released Library Specifications.}
Upon acceptance, we will share a library of code. The library allows using the metrics used in this paper as well as easily defining new ones. It reports MuLER variants as well as discrepancy breakdowns (\S\ref{sec:hallucination}).

\subsection{Datasets} \label{subsec:data}

\paragraph{WMT.} We use the official submissions and references from \href{https://www.statmt.org/wmt21/results.html}{WMT} $2014$-$2020$ news translation task \citep{bojar-etal-2014-findings, bojar-etal-2015-findings, bojar-etal-2016-findings, bojar-etal-2017-findings, bojar-etal-2018-findings, barrault-etal-2019-findings, Barrault2020FindingsOT}.
We use all language pairs in each year with English as a target language. 

\paragraph{Gender.} We make use of the \href{https://github.com/rudinger/winogender-schemas}{WinoGender} dataset \citep{rudinger-EtAl:2018:N18} where each sentence has a variation of male, female and neutral (App.~\S\ref{ap:para}).

\paragraph{Paraphraes.} We use the Minimal Paraphrase pairs corpus by \citet{patel-etal-2022-neurons}. It contains parallel corpora with two syntactic variation types: active versus passive sentences and adverbial clause versus noun phrases. The changes to the sentences are minimal, specifically, the semantic meaning remains identical. See App~\ref{ap:para} for more details.

\section{Experiments with Naturalistic Data}\label{sec:analysis}

\subsection{Single Model Analysis} \label{subsec:model_analysis}

A key point of MuLER is the ability to compare the performance of various features on a single model. 

\begin{figure}
\centering
\includegraphics[width=\columnwidth]{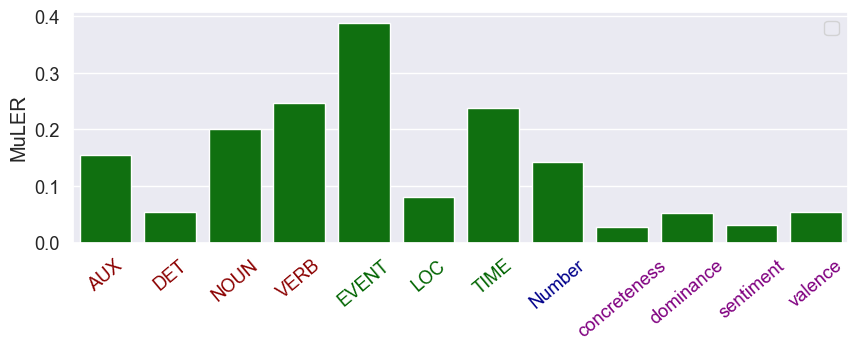}
\caption{Standard MuLER report. Chinese-English for a subset of features. The Newstest2020 dataset. Submission {\it Huoshan Translate.919}.}

\label{fig:muler_report}
\end{figure}

Such an analysis can reveal the system's strengths and weaknesses and potentially lead to a targeted development effort on specific features, or be used for debugging purposes. It enables the user to decide where to invest his efforts and allows for a more scientifically-oriented investigation of the results. 
Fig.~\ref{fig:muler_report} shows a standard MuLER report for two systems. 
\subsection{Comparison Across Systems}\label{subsec:comparing_models}

We compare WMT systems through years, architectures and performance patterns.

\begin{figure*}[tbh]  
      \centering
    \begin{subfigure}{.99\textwidth}
      \includegraphics[width=1\linewidth]{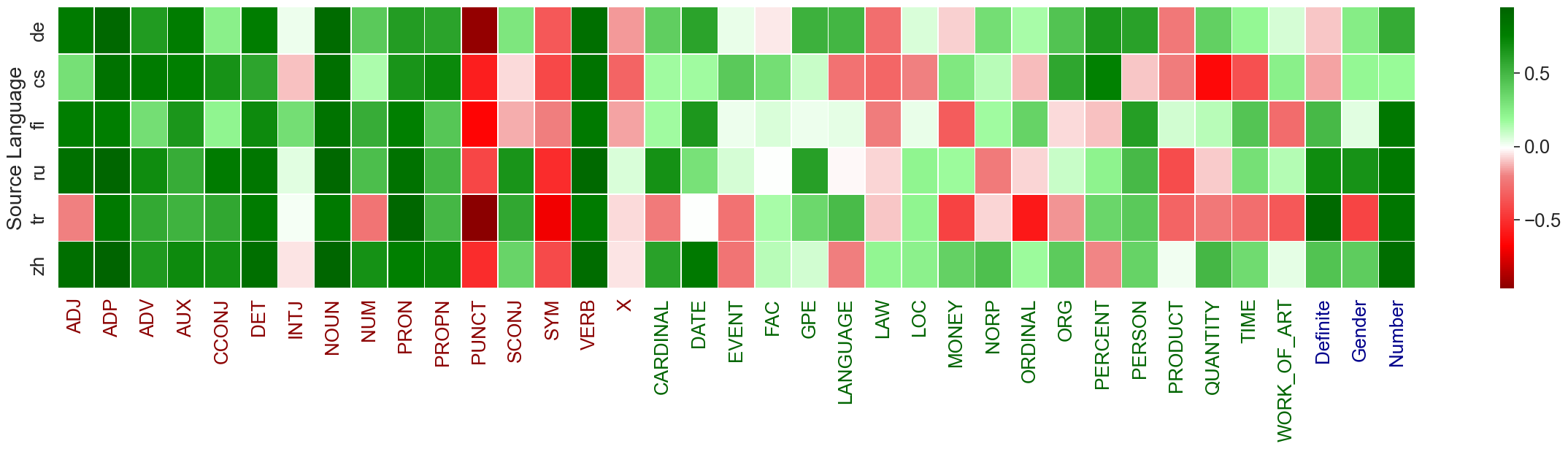}
    
    \end{subfigure}
    \caption{Similarity of Measures. Correlation between BLEU and -MuLER per feature (column) and source language (row). Positive values suggest better systems by BLEU better translate the feature.}
    \label{fig:bleu_sim}
\end{figure*}

\paragraph{MuLER Similarity to Other Measures.}  We compute Pearson correlation between negative MuLER scores and BLEU, for every source language, over all submissions of WMT ($2014-2017$). We use negative MuLER so that high correlation means improvements in both performance measures (e.g., BLEU and MuLER), as reference and candidate similarity is indicated by high BLEU but low MuLER. Fig.~\ref{fig:bleu_sim} shows that BLEU and MuLER are not always correlated. We see that arousal, concreteness, dominance, sentiment and valence scores are in high agreement between MuLER and BLEU. However, some features, e.g., most of the named entity types, are not. This suggests that overall BLEU improvements do not necessarily mean better named entity translations.

We also see that different languages behave differently with respect to the type of features for which MuLER and BLEU are highly correlated. For example, in Chinese, BLEU is more correlated with MuLER, over many different POS tags. This could be explained by differences in the structure of the languages (e.g., syntax). A possible explanation might be that Chinese is simpler to translate in terms of overlapping unigrams (i.e., when syntax is ignored).
We do the same analysis comparing MuLER to indices-BLEU (BLEU over the indices in which the feature appears both in the reference and the output) and their $max(R,C)-min(R,C)$ term. We get similar results (see App. \ref{app:fig:bleu_sim}).

\begin{figure}[t]
\centering
\includegraphics[width=\columnwidth]{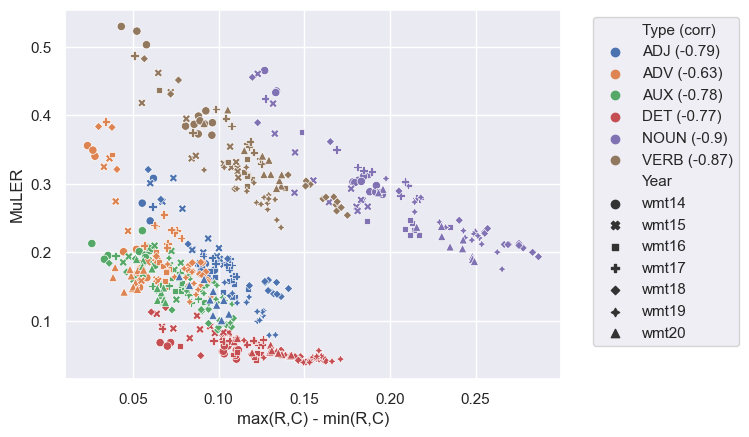}
\caption{MuLER vs. max(R,C) minus min(R,C) calculated on selected POS-tags. All submissions to WMT ($2014-2020$) for German-English. Next to each POS-tag is the correlation between all x-axis and y-axis values for the POS tag.}
\label{fig:lang_scat_pos_deen}
\end{figure}

\begin{figure}[t]
\centering
\includegraphics[width=\columnwidth]{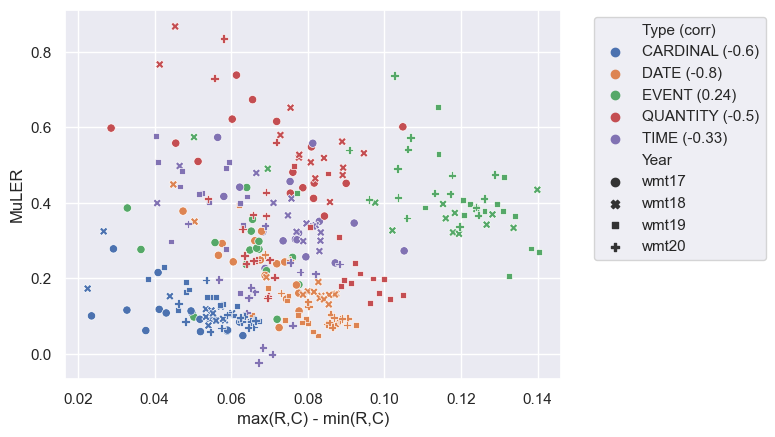}
\caption{MuLER vs. max minus min calculated on named entities. All submissions to WMT ($2017-2020$) for Chinese-English. Next to each entity is the correlation between x-axis and y-axis values for the entity.} 

\label{fig:lang_scat_ner_zhen}
\end{figure}


\begin{figure}[t]
\centering
\includegraphics[width=\columnwidth]{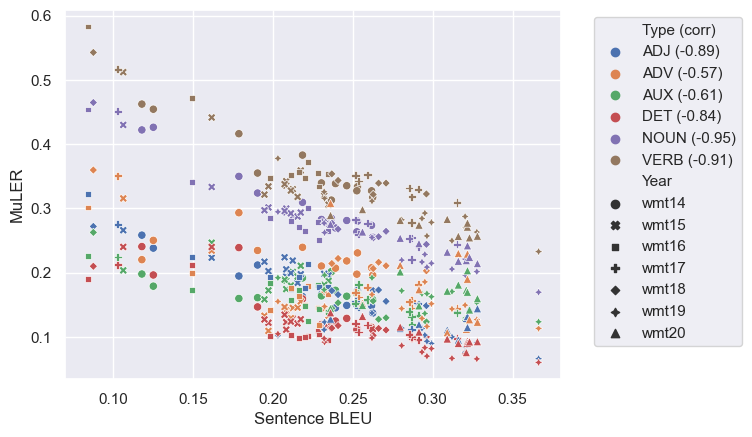}
\caption{POS-tag MuLER vs. BLEU. All submissions to WMT ($2014-2020$) for Russian-English. Next to each POS-tag is the correlation between all x-axis and y-axis values for the POS-tag.}
\label{fig:lang_scat_pos_deen_bleu_MuLER}
\end{figure}

\paragraph{Systems Over Time.} We compare WMT systems (see \S\ref{subsec:data}) from different years and language pairs with MuLER.

Overall, there is a consistent trend  (see Figs.~\ref{fig:lang_scat_pos_deen},\ref{fig:lang_scat_ner_zhen},\ref{fig:lang_scat_pos_deen_bleu_MuLER}): as BLEU improves, MuLER improves.  

However, this trend is not uniform across all features. For certain phenomena, improvement is not consistent with system quality. This is shown by a near-zero or positive correlation between MuLER and the $max(R,C)-min(R,C)$ term (indicative of the system's performance on the sentences containing $f$).

Surprisingly, we find that nouns and verbs are among the hardest POS tags to translate (Fig.~\ref{fig:lang_scat_pos_deen}). On the face of it, this is unexpected, as they account for the most frequent POS tokens in training. Potentially, being open class makes them harder, nouns are common, but each noun by itself is rare. This may also explain why determiners that are frequent are easy and why adverbs are harder than the more frequent auxiliary.
Similar trends are presented when comparing MuLER to the total BLEU score of the systems (Fig.~\ref{fig:lang_scat_pos_deen_bleu_MuLER}).

\begin{figure}
\centering
\includegraphics[width=\columnwidth]{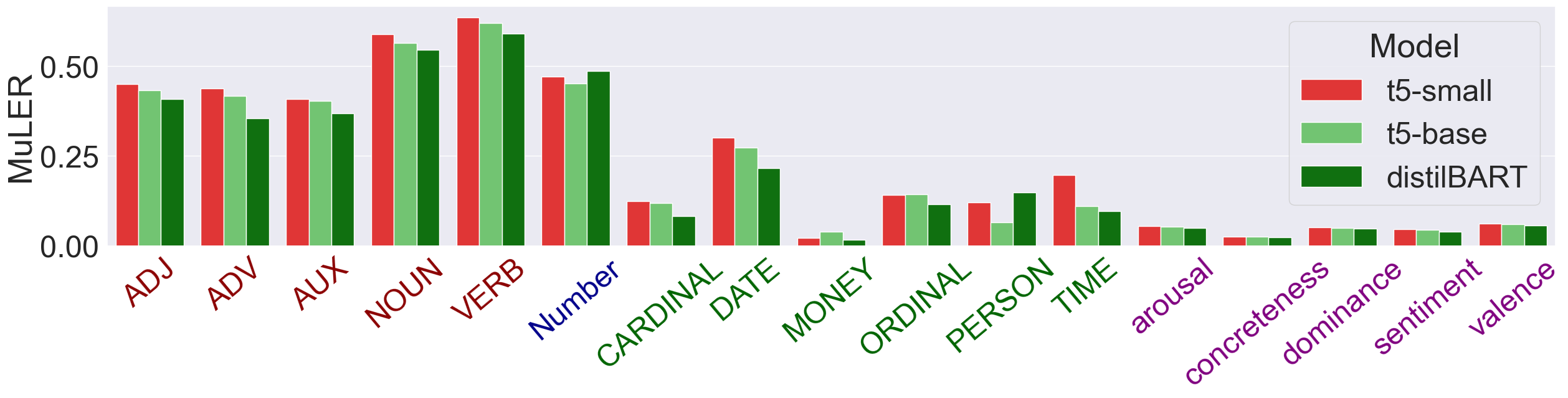}
\caption{MuLER for summarization. MuLER score is calculated for various features, under ROUGE. We compare $3$ models; t5 small, t5 base and distill BART.}

\label{fig:sum_t5}
\end{figure}

\begin{table*}[!t]
\centering
\footnotesize
\renewcommand{\tabcolsep}{0.15cm}
\begin{tabular}{p{0.7cm}p{1.2cm}p{1.cm}p{1.3cm}p{3.0cm}p{3.0cm}p{3.0cm}}
\toprule
\textbf{Year} & 
\textbf{Language pair} & 
\textbf{Feature type} & 
\textbf{Feature} & 
\textbf{Reference} &
\textbf{System A} &
\textbf{System B} \\

\midrule

2020 & ru-en & POS & AUX & "This \greenemph{is} heavy oil. & "This \greenemph{is} thick oil. & "It'\redemph{s} thick oil. \\

\midrule

2019 & fi-en & NER & LOC & Daytime temperatures are between 7 and 12 degrees Celsius, but cooler in \greenemph{Northern Lapland}. & Daytime temperatures are between + 7 and + 12 degrees, it's cooler in \greenemph{northern Lapland}. & Daytime temperatures are between + 7 and + 12 degrees, the North is cooler \redemph{Lapland}. \\
\midrule

2018 & tr-en & POS & ORDINAL & \greenemph{Thirdly}, technology is developing very fast. & \greenemph{Thirdly}, technology is evolving rapidly. & \redemph{Third}, technology is evolving rapidly. \\

\midrule

2018 & tr-en & POS & ADJ & \greenemph{Single} digit inflation & Inflation is \greenemph{single} digits & Inflation is the \redemph{only} household \\

\midrule

2018 & tr-en & POS & ADJ & Clearly, the murders have a \greenemph{chilling} effect. & The killings clearly had a \greenemph{chilling} effect. & The killings have clearly had a \redemph{cold} shower effect. \\

\bottomrule
\end{tabular}

\caption{Example sentences from WMT's submissions. System A has a lower MuLER score than system B. We indicate whether the chosen feature is \greenemph{consistent} or \redemph{inconsistent} with the reference.}
\label{table_qualitative}
\end{table*}

\subsection{Manual Analysis} 
\label{subsec:manual_analysis}
To verify the effectiveness of MuLER, we perform manual analysis and compare pairs of systems that are roughly equal in their overall performance (under BLEU), but greatly differ on a given feature $f$ (under MuLER). We compare $5$ pairs of systems and a total of $201$ sentences (App.~\S\ref{app:tab:manual_analysis}).

We consistently see that systems with lower MuLER scores (i.e., better performance) translate feature $f$ better (see Table \ref{table_qualitative}). This means that the neighborhood of $f$ in the candidate sentence is more similar to the reference, not only the masked span itself.
Interestingly, we encounter many cases in which the span of $f$ is the same in the reference and both candidates, but the overall translation (i.e., the neighborhood) is better in the one with the lower MuLER.
Table~\ref{app:tab:manual_analysis} shows  that out of $97$ sentences where quality differs, the system MuLER predicts to be better, indeed translates better in $91.3\%$ of the sentences.   


\begin{table*}[!t]
\centering
\footnotesize
\renewcommand{\tabcolsep}{0.15cm}
\begin{tabular}{ccccc|ccc}
\toprule
\multicolumn{5}{c|}{\textbf{synthetic features}} & \multicolumn{3}{c}{\textbf{features}} \\
\cmidrule(lr){1-5}\cmidrule(lr){6-8}
\thead{\textbf{average proportion} \\ \textbf{(reference)}} & 
\thead{\textbf{average proportion} \\ \textbf{(output)}} &
\thead{\textbf{average} \\ \textbf{MuLER}} & 
\thead{\textbf{variance} \\ \textbf{MuLER}} & 
\thead{\textbf{std} \\ \textbf{MuLER}} & 
\textbf{feature} & 
\thead{\textbf{average} \\ \textbf{proportion}} & 
\textbf{MuLER} \\

\midrule
\greenmark{0.22} & 
\greenmark{0.22} & 
\redmark{0.44} & 
4.09e-04 & 
0.01 & 
NOUN & 
\greenmark{0.22} & 
\redmark{0.26} \\

\midrule

\greenmark{0.15} & 
\greenmark{0.15} & 
\redmark{0.22} & 
2.24e-04 & 
0.01 & 
VERB & 
\greenmark{0.12} & 
\redmark{0.29} \\

\midrule

\greenmark{0.11} & 
\greenmark{0.11} & 
\redmark{0.21} & 
6.04e-04 & 
0.03 & 
PROPN & 
\greenmark{0.09} & 
\redmark{0.07} \\

\midrule

\greenmark{0.07} & 
\greenmark{0.07} & 
\redmark{0.21} & 
2.53e-04 & 
0.02 & 
PRON & 
\greenmark{0.07} & 
\redmark{0.16} \\

\bottomrule
\end{tabular}
\caption{Specificity of MuLER. Comparison of \redmark{MuLER} for synthetic features ("average MuLER") with real features ("MuLER"). The two leftmost columns are the \greenmark{average proportion} of the synthetic features in the reference and output. The "average proportion" column indicates the average frequency of the features (e.g, NOUN/VERB) in the reference and the output (as described in \S\ref{sec:validation}). WMT $2019$ submission; "online-G.0" for German-English.}
\label{tab:synthetic_features}
\end{table*}


\subsection{MuLER with ROUGE: Summarization}

We compute MuLER on $3$ summarization models (App.~\S\ref{sec:summarization_setup}) and various features. Fig.~\ref{fig:sum_t5} shows a standard MuLER report, computed under the ROUGE metric. We see that strengths and weaknesses vary between the different systems. Moreover, we see that the concreteness score is always lower than the other scores provided by the sentence scorers (i.e, valence, dominance, arousal and sentiment). Inherently, we expect summarization outputs to be concrete, as compressing the text is often achieved by simplification. This is indeed revealed by MuLER.

\subsection{MuLER with LM-based Metrics}
To validate that MuLER could be easily adapted to LM-based metrics, in addition to BLEU, we perform our analysis for the task of MT, also with BERTScore (\cite{bert-score}). We randomly choose $5$ systems from WMT-$2020$ for Chinese-English.  Preliminary experiments show that MuLER can be straightforwardly extended (App.~\S\ref{ap:bertscore}) to such metrics.

\subsection{Paraphrases and Gender} \label{subsec:paraphrases_gender}
\begin{figure}[tbh]
\centering
    \begin{subfigure}{.22\textwidth}
      \centering
      \includegraphics[width=1\linewidth]{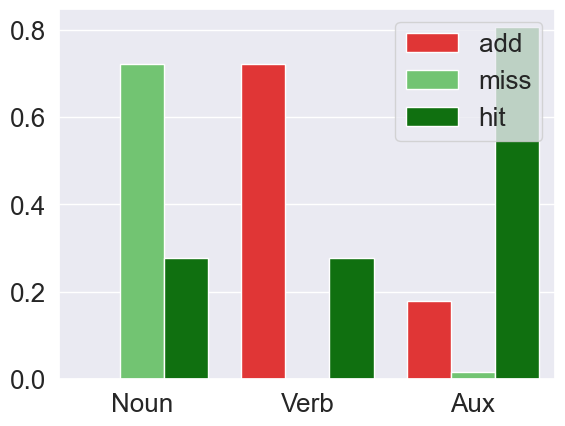}
      \caption{Clause / Noun Phrase}
      \label{fig:passive2active_split}
    \end{subfigure}
    \begin{subfigure}{.22\textwidth}
      \centering
      \includegraphics[width=1\linewidth]{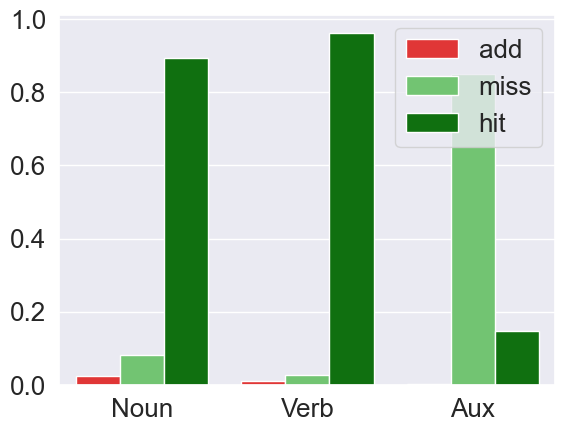}
      \caption{Active / Passive}
      \label{fig:active2passive_split}
    \end{subfigure}
    \caption{Discrepancy breakdown of verbs, nouns and auxiliaries for minimal syntactic paraphrases.}
    \label{fig:hall}
\end{figure}

We apply MuLER to special cases to demonstrate its usefulness.
\paragraph{Minimal Paraphrases.} We compare Minimal Paraphrases (\S\ref{subsec:data}, App.~\S\ref{ap:para}) as if they were an output and reference. Evidently, the discrepancy breakdown identifies phrasing differences (see Fig.~\ref{fig:hall}). Adverbial clause sentences have more verbs, while noun phrases have more nouns and thus their miss and hit scores complement each other. The scores also recognize voice changes from active to passive; these require additional auxiliaries while keeping the same verbs and nouns.

\paragraph{WinoGender.} Gender choice is critical for many applications. We compare sentences which differ only by gender (\S\ref{subsec:data}, App.~\S\ref{ap:para}) as if they were an output and reference.
Where sentences with different gender receive a high BLEU score (0.8), the gender feature of MuLER is 1.0 -- representing the  perfect inability of the systems to translate the correct gender. This shows the strength of MuLER over bottom-line metrics (e.g, BLEU) as it reveals the performance on a specific dimension (gender).

\section{Validation Experiments} \label{sec:validation} 

In this section, we perform various synthetic experiments to check the validity of MuLER. For a given feature $f$, let $\mathcal{F}$ be the set of words tagged as $f$ (e.g., nouns) under $\tau$, and $\alpha \in [0,1]$.


\begin{table*}[t!] 
\centering
\footnotesize
\begin{tabular}{ccccccccc}
\toprule
\textbf{system} &
\multicolumn{2}{c}{\textbf{50\% abl-MuLER}} &
\multicolumn{2}{c}{\textbf{100\% abl-MuLER}} &
\multicolumn{2}{c}{\textbf{50\% MuLER}} &
\multicolumn{2}{c}{\textbf{100\% MuLER}} \\
\cmidrule(lr){2-3}\cmidrule(lr){4-5}\cmidrule(lr){6-7}\cmidrule(lr){8-9}
 & noun & verb & noun & verb & noun & verb & noun & verb \\
\midrule
Facebook\_FAIR.6750 &
0.021 & 0.018 & 0.054 & 0.034 & 0.203 & 0.320 & 0.267 & 0.391 \\ 
\midrule
online-A &
0.023 & 0.017 & 0.055 & 0.036 & 0.229 & 0.357 & 0.295 & 0.432 \\
\midrule
UCAM.6461. &
0.023 & 0.017 & 0.054 & 0.035 & 0.220 & 0.328 & 0.279 & 0.405 \\
\bottomrule
\end{tabular}
\caption{Robustness to Feature Frequency. Presented here are $3$ submissions from WMT $2019$, translation from German to English (see Table \ref{app:tab:frequency} for more results). We compare between MuLER and abl-MuLER (MuLER's numerator -- an ablated version of MuLER) with $50 \% / 100 \%$ of nouns/verbs masked.
\label{tab:frequency}}
\end{table*}


\paragraph{Range and Monotonicity of MuLER.} 
We expect MuLER to fall in the interval $[\sigma(min(R,C)),\sigma(max(R,C))]$ and to improve as the quality of translation on the feature $f$ improves (monotonicity). That is, if a system outputs the right translation for $\alpha$ cases of $\mathcal{F}$ (and wrong on $1-\alpha$ cases accordingly), then we expect $MuLER(R,C) \approx \alpha (\sigma(max(R,C))-\sigma(min(R,C)))$.

We support this claim using synthetic data experiments. We define a hybrid version of MuLER using a combination of oracle (O) and anti-oracle (AO) masking strategies (\S\ref{subsec:tagger}). 
We split $\mathcal{F}$ into two sets roughly containing $\alpha$ and $1 - \alpha$ of its elements, by partitioning according to sorted first letter. That is, we choose $\eta$ to be the first letter in the English Alphabet for which the set of all words in $\mathcal{F}$ that start with a-$\eta$ is of size $\geq \alpha \mathcal{F}$. We split $\mathcal{F}$ to $2$ sets; one containing all words that start with the letter a-$\eta$, and its complement.
We mask $\alpha$ of the occurrences of $f$ using AO-strategy, and the rest using O-strategy, both in the reference and the candidate.
This construction emulates a range of systems that improve on $f$ as a function of $\alpha$. 

Tables \ref{tab:hybrid_table},\ref{app:tab:hybrid_muler4}, \ref{app:tab:hybrid_muler3} show that this hybrid score is indeed always located according to $X$ in the interval  $[ \textnormal{min}_{\sigma}(R,C), \textnormal{max}_{\sigma}(R,C) ]$ (e.g., if $X=2$ then it's in the middle of the interval). 

\begin{table*}[t]
\centering
\footnotesize
\begin{tabular}{p{0.3cm}p{0.75cm}p{2.7cm}p{0.9cm}p{0.35cm}p{0.35cm}p{0.35cm}p{0.35cm}p{0.35cm}p{0.35cm}p{0.35cm}p{0.35cm}p{0.35cm}p{0.35cm}}
\toprule
\textbf{year} &
\textbf{langs} &
\textbf{submission} &
\textbf{system bleu} & 
\multicolumn{2}{c}{\textbf{bleu indices}} &
\multicolumn{2}{c}{\textbf{MuLER}} &
\multicolumn{2}{c}{\textbf{O}} & 
\multicolumn{2}{c}{\textbf{AO}} &
\multicolumn{2}{c}{\textbf{hybrid}} \\
\cmidrule(lr){5-6}\cmidrule(lr){7-8}\cmidrule(lr){9-10}\cmidrule(lr){11-12}\cmidrule(lr){13-14}
 & & & & n & v & n & v & n & v & n & v &
 n & v \\

\midrule

20 & 
de-en & 
newstest2020.de-en.OPPO.1360 & 
0.39 & 
0.41 & 
0.41 & 
0.18 & 
0.29 & 
0.45 & 
0.45 & 
0.21 & 
0.32 & 
0.33 & 
0.38 \\

\midrule

18 & 
ru-en & 
\thead[l]{newstest2018.Alibaba. \\ 5720.ru-en} & 
0.30 & 
0.30 & 
0.30 & 
0.24 & 
0.32 & 
0.35 & 
0.34 & 
0.14 & 
0.21 & 
0.24 & 
0.27 \\

\midrule

15 & 
fi-en & 
\thead[l]{newstest2015.uedin- \\ syntax.4006.fi-en} & 
0.12 & 
0.12 & 
0.13 & 
0.38 & 
0.39 & 
0.17 & 
0.16 & 
0.05 & 
0.08 & 
0.10 & 
0.12 \\

\bottomrule
\end{tabular}
\caption{Range and Monotonicity of MuLER. MuLER scores on nouns ("n") and verbs ("v") in $5$ randomly chosen systems from WMT. Oracle ("O") and Anti-Oracle ("AO") masking strategies vs. hybrid masking strategy (described in \S\ref{sec:validation}) at $50-50$ split ($50 \%$ of noun/verb is masked with O-strategy, and the rest with AO-strategy).  \label{tab:hybrid_table}}

\label{table min pairs}
\label{tab:main_examples}
\end{table*}

\paragraph{Specificity of MuLER.}

We set to verify that MuLER is not sensitive to random features in the text. We expect that features that appear in random subsets of the text with the same frequency  will have roughly the same score.
To verify this, we create synthetic features with the same frequency in $\mathcal{F}$ as real ones (e.g, nouns/verbs) and compute MuLER over them. 

Let $U$ be the unique list of words in the union of $R$ and $C$. For $1 \leq j \leq 1000$: we split $U$ to $p$ equally sized groups $\{U_1,...,U_p \}$ (we ignore the remainder). Indeed, as seen in Table \ref{tab:synthetic_features}, the average proportion of $U_i$ in $R$ and $C$ is roughly the same.
For $1 \leq  i \leq p$ we compute $MuLER(R,C)$ by masking only the words in $U_i$ (both in $R$ and $C$). At each run we have $p$ scores $\{(m_1,...,m_p)_j \}_{j=1}^{1000}$ from which we choose one randomly.  In total, we get $1000$ scores: $M = \{ \tilde{m}_1,...,\tilde{m}_{1000} \}$. We compute the variance and standard deviation for $M$ (see Table \ref{tab:synthetic_features}). We find that the variance and std are around zero across values of $p$, for $p \in \{ 2,...,6\}$ (see App.~\S\ref{app:tab:synthetic_features}). Meaning, MuLER is not specified to random phenomena. Moreover, the results are different compared to \emph{real} linguistic phenomena with the same frequency (e.g, nouns/verbs, see Table \ref{tab:synthetic_features}). 
These findings suggest that MuLER is not sensitive to variation that does not reflect variation in quality.

\paragraph{Robustness to Feature Frequency.} We start by validating that MuLER score is less sensitive to the frequency of $f$. 

We split $\mathcal{F}$ into two sets roughly containing $\alpha$ and $1 - \alpha$ of its elements, by partitioning according to sorted first letter (as explained before). We then mask $\alpha$ of $\mathcal{F}$ and ignore the rest of the instances. This allows us to test MuLER on a feature with similar performance (a random sample of the original feature) but different frequency, namely $\alpha$ frequency of the feature $f$ across $\mathcal{F}$ (this is not true when doing the split at the sentence-level). We see in Table~\ref{tab:frequency} that MuLER is robust to changes in frequencies (of nouns and verbs), compared to abl-MuLER -- an ablated version of MuLER which is defined as MuLER's numerator. This holds across various frequencies and features (see Table \ref{app:tab:frequency}). This suggests that MuLER is a more suitable score for measuring system performance and that its signal is not due to the frequency of the feature (it may play a role, but not a central one).  


\section{Related Work}\label{sec:background}

Automatic metrics are useful to assess systems and we base our work on them (see \S\ref{subsec:eval_metrics}). Other lines of work study a specific property and propose evaluation measures for it. For example, addressing hallucinations \citep{kryscinski2020evaluating} or measuring grammaticality \citep{vadlapudi2010automated}. We share the aspiration to a more fine-grained form of evaluation with these works.

There are methods for analyzing performance in a more fine-grained manner. For example, evaluation with minimal changes to the input \citep{warstadt2020blimp} and challenge sets \citep{macketanz2018fine,emelin-sennrich-2021-wino}. 

Few methods highlight patterns rather than predefined properties, by contrasting texts (e.g. reference and output) \citep{Gralinski2019GEvalTF, lertvittayakumjorn2021grasp}. In a sense, MuLER stands in the middle between those, it highlights a closed set of traits, but it is extendable.

\section{Conclusion}

We presented a novel methodology (MuLER) to decompose any reference-based score into its fine-grained components, making it possible to obtain a detailed picture of text generation systems' performance, instead of a bottom-line score. 

MuLER filters and dissects naturalistic data to highlight phenomena in the generated text. We validated MuLER using a set of synthetic experiments (\S\ref{sec:validation}). 

Applying MuLER to off-the-shelf systems shows (\S\ref{sec:analysis}) that different systems' strengths and weaknesses are varied, even when their overall performance is alike, and detect interesting trends over the years. 
Our work creates an avenue for further research into more fine-grained evaluation metrics and provides a tool to understand system behaviour. In future work, we plan to extend MuLER to more complex features such as long-distance syntactic dependencies.

\section*{Limitations}

Among MuLER appealing traits is its reliance on existing, accepted and easily changed components. It also counts as its limitation, where the base metric is invariant to a trait, MuLER would also be, where masking tagging or scoring is not available (e.g. in endangered languages) the features would not be possible to extract. In general, detecting a feature (e.g. POS tag) is usually harder than evaluating the quality of its generation, MuLER makes this evaluation more accessible.

By definition, MuLER is as good as the tagger that is used to detect a feature of choice. While there is a potential for noise in the process, the taggers used in this paper are known to work well and are indeed vastly used.

We showcase MuLER on BLEU and ROUGE as they are still among the most widely adopted metrics in their respective tasks. The concept of MuLER can be straightforwardly extended to LM-based metrics and we intend to explore it in future work. For now, we shared initial results on BERTScore suggesting this is indeed the case.

For some validations, we use synthetic experiments, that make a well-controlled experiment, but sometimes lack some characteristics of natural data.
Overall, we try to evaluate intrinsically, extrinsically by use cases, manually and synthetically to present a full view where the whole is greater than the sum of its parts.

Although we use MuLER to compare between models, it is not clear whether such a comparison is interesting for systems with overall very different performance; if one system's overall performance is very low, then even if it somehow translates a specific feature well, the quality of its output is bad. However, comparing systems with overall similar performance is the more common use case and hence useful; for example, when choosing between systems with top performance to perform a task or for analyzing the differences between systems.

\section*{Acknowledgements}
 
This work was supported
in part by the Azrieli Fellowship, the Vatat scholarship, the Israel Science Foundation (grant
no. 2424/21), and by the Applied Research in
Academia Program of the Israel Innovation Authority. 

\bibliography{anthology,custom}
\bibliographystyle{acl_natbib}

\appendix

\section{Scorers used}\label{ap:sec:scorers}
In this section, we elaborate on the scorers' use and their origin.
\paragraph{Sentiment.} Sentiment Analysis is the process of determining whether a piece of text is positive, negative or neutral. We follow the method of \citet{Khoo2018LexiconbasedSA} that relies on
per word score and a rule-based combination (mainly dealing with negation). The method was shown to outperform other lexicons and to work well without the need for neural networks.
We selected this method as it strikes a good balance between accuracy and running time. We defer the application of neural metrics to future work.

We consider 4 token-level scores which we aggregate into a sentence score by averaging. We ignore words that do not appear in the lexicons.

\paragraph{Concreteness.} The Concreteness rating of a word represents to which extent a word is concrete, how perceptible is it. For example, a fruit is less concrete than a banana and tomorrow is more concrete than sometime.
The lexicon \citep{Brysbaert2014ConcretenessRF} contains 40K lemmas each with a concreteness score.  

\paragraph{Valence Arousal and Dominance.} In psychology, it is common to discuss three characteristics in how we perceive others (e.g., in recognizing faces \citep{jones2021world}): valence (pleasure vs.
displeasure), arousal (active vs. passive), and dominance (dominant vs. submissive). These were shown to be mostly independent directions of word meaning \citep{osgood1957measurement, russell1980circumplex, russell2003core}. The lexicon \citep{mohammad2018obtaining} contains 20K words and their respective scores for each of those axes.

\section{Summarization}\label{sec:summarization_setup} 
We compare T5-base \citep{raffel2020exploring}, T5-small and distillbart \citep{shleifer2020pre,lewis2020bart} models on the CNN Daily Mail summarization dataset \citep{Nallapati2016Abstractive}.

We use models from the \href{https://huggingface.co/models}{HuggingFace} model hub.

DistillBart-"sshleifer/distilbart-cnn-12-6"
and T5-"t5-base" and "T5-small"

\section{LM-based Metrics}
\label{ap:bertscore}
We perform preliminary experiments using \href{https://github.com/Tiiiger/bert_score}{BERTScore}, which is a language-model (LM) based metric for measuring the quality of generation tasks. We use it together with "bert-based-uncased" model. In order to adapt BERTScore to MuLER, we perform alterations to the similarity matrix of the reference and candidate embeddings, that is calculated during the score's computation. 
To compute ${\greenemphmath{\textnormal{max}_{\sigma}(R,C)}}$, after the similarity matrix between the un-masked reference and un-masked candidate is computed, we set the $ij$-th entry to be $1$ if both the $i$-th word in the reference and the $j$-th word in the candidate is masked (if the masked word is split to multiple tokens by the BERT tokenizer, we set the corresponding entry in the similarity matrix to be $1$ for each of them). To compute ${\redemphmath{\textnormal{min}_{\sigma}(R,C)}}$, after the similarity matrix between the un-masked reference and un-masked candidate is computed, we set the $i$-th row to be zeroes if the $i$-th word in the reference is masked, and the $j$-th column to be zeroes if the $j$-th word in the candidate is masked. Indeed, in this setting we also get that $\greenemphmath{\textnormal{min}_{\sigma}(R,C)} > \redemphmath{\textnormal{min}_{\sigma}(R,C)}$ (this is true for $1000$ randomly sampled sentences from the submissions we analyzed).
We randomly sampled $5$ submissions to WMT-2020 for Chinese-English (Tencent\_Translation.1249, Online-B.1605, DeepMind.381, Huoshan\_Translate.919 and OPPO.1422). Similar trends to the results obtained by MuLER with BLEU are exhibited.  

\begin{figure*}[tbh]
    \begin{center}
      \includegraphics[width=.99\textwidth]{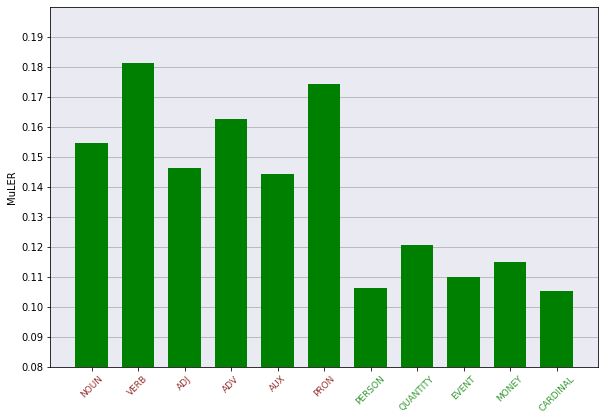}
    \end{center}
    \caption{Standard MuLER Report with BERTScore. Chinese-English for a subset of features. The Newstest2020 dataset. Submission "Huoshan Translate.919". MuLER computed with BERTScore.}
    \label{ap:fig:bertscore_single_report}
\end{figure*}

\section{Data}
\label{app:sec:data}
We provide the complete MuLER database containing the results for WMT submissions ($2014-2020$) on all features (see \ref{subsec:features}) in the supplementary materials (App.~\S\ref{app:supp}). We will release it together with our code upon acceptance.

\section{Supplementary Materials}
\label{app:supp}
The complete MuLER database (scores for all WMT's submissions ($2014-2020$)) and the tagged manual analysis are in the supplementary materials submitted with the paper. 

\subsection{Minimal Paraphrases}\label{ap:para}
Minimal Paraphrases dataset \citep{patel-etal-2022-neurons} contains 1169 active-passive pairs and 114 clause-noun phrase pairs. Examples are in table \ref{tab:data_examples}.
\begin{table*}[ht]
\begin{small}
\begin{center}
\begin{tabular}{@{}lll@{}}
\toprule & Source & Paraphrased \\ \midrule
\begin{tabular}[c]{@{}l@{}}Active Voice$\rightarrow$ Passive Voice\end{tabular}   & \textit{She \textbf{took} the book}                                                                & \textit{The book \textbf{was taken} by her}                                                        \vspace{0.1cm}\\
\begin{tabular}[c]{@{}l@{}}Adverbial Clause$\rightarrow$ Noun Phrase\end{tabular} & \textit{\begin{tabular}[c]{@{}l@{}}The party died down before \textbf{she arrived}\end{tabular}} & \textit{\begin{tabular}[c]{@{}l@{}}The party died down before \textbf{her arrival}\end{tabular}} \\ \bottomrule
\end{tabular}
\end{center}
\vspace{-0.25cm}
\caption{Examples of minimal paraphrases}
\label{tab:data_examples}
\end{small}
\end{table*}


\begin{table*}[ht]
\begin{small}
\begin{center}
\begin{tabular}{@{}lc@{}}
\textit{The technician told the customer that \textbf{she} could pay with cash. }                       \\
\textit{The technician told the customer that \textbf{he} could pay with cash.}                         \\ \midrule
\textit{The supervisor gave the employee feedback on \textbf{her} stellar performance. }                \\
\textit{The supervisor gave the employee feedback on \textbf{his} stellar performance. }                \\ \midrule
\textit{The librarian helped the child pick out a book because \textbf{she} did not know what to read.} \\
\textit{The librarian helped the child pick out a book because \textbf{he} did not know what to read. }
\end{tabular}
\caption{Female-Male pairs from the WinoGender dataset}
\label{tab:winogender}
\end{center}
\end{small}
\end{table*}


\FloatBarrier

\subsection{WinoGender}\label{ap:gender}
WinoGender \citep{rudinger-EtAl:2018:N18} cosists of sentences that differ only by the gender of one pronoun in the sentence, see examples in Table~\ref{tab:winogender}.

\section{Manual Analysis}
\label{ap:sec:manual_analysis}
We perform a small-scale manual analysis to validate MuLER does indicate the quality of performance on a certain feature. We chose $5$ systems from different years and language pairs (see Table \ref{app:tab:manual_analysis} for full details). We compare pairs of systems that are roughly equal in their overall performance (under BLEU), but greatly differ on a given feature $f$, under MuLER (see \S\ref{subsec:manual_analysis}). One of the authors annotated the data. For every pair of submissions, the data was shuffled such that the sentences were side by side without knowing in advance which is the better system.

\section{Negative MuLER}

Intuitively, we expect to always gain by masking a certain proportion of a given feature in the text (i.e, positive MuLER score). However, there are edge cases in which $max(R,C) - BLEU(R,C)$ is negative. It can be due to a mistake of the tagger or the sentence structure (for example, a word in the reference that is a noun is used in the candidate as a verb, etc.). In table \ref{app:tab:neg_score} we present examples for such cases.

\begin{table*}[!t]
\centering
\footnotesize
\renewcommand{\tabcolsep}{0.15cm}
\begin{tabular}{p{3.4cm}p{3.4cm}p{3.4cm}p{3.4cm}}
\toprule

\textbf{reference} &
\textbf{masked reference} &
\textbf{output} &
\textbf{masked output} \\

\midrule

\greenemph{Nitromethane} is being used for \greenemph{example} in \greenemph{drag} \greenemph{racing}. &
\greenemph{NOUN} is being used for \greenemph{NOUN} in \greenemph{NOUN} \greenemph{NOUN}. &
\redemph{Nitromethane} is used, for \redemph{example}, drag \redemph{racing}. &
\redemph{NOUN} is used, for \redemph{NOUN}, drag \redemph{NOUN}. \\

\midrule

The \greenemph{film} will premiere in Finland in September 2015. &
The \greenemph{NOUN} will premiere in Finland in September 2015. &
The \redemph{film} will have its Finnish \redemph{premiere} in September 2015. &
The \redemph{NOUN} will have its Finnish \redemph{NOUN} in September 2015. \\

\midrule

Its unpredictability unsettled \greenemph{people}'s \greenemph{nerves}. &
Its unpredictability unsettled \greenemph{NOUN}'s \greenemph{NOUN}. &
Its \redemph{unpredictability} made \redemph{people} nervous. &
Its \redemph{NOUN} made \redemph{NOUN} nervous. \\

\midrule

Our whole \greenemph{house} moved, we were trembling with \greenemph{fear}. &
Our whole \greenemph{NOUN} moved, we were trembling with \greenemph{NOUN}. &
We need the \redemph{whole} of our \redemph{house} moved: \redemph{vapisimme} \redemph{fear}. &
We need the \redemph{NOUN} of our \redemph{NOUN} moved: \redemph{NOUN} \redemph{NOUN}. \\

\bottomrule
\end{tabular}
\caption{Negative MuLER.}
\label{app:tab:neg_score}
\end{table*}

\FloatBarrier

\section{graphs}
\label{ap:graphs}

We supply here multiple graphs that were mentioned in the text. The rest of the analysis graphs could be found in the supplementary files.

\begin{table*}[!t]
\centering
\footnotesize
\renewcommand{\tabcolsep}{0.15cm}
\begin{tabular}{p{0.7cm}p{1.2cm}p{1.3cm}p{1.2cm}p{7.5em}p{3.0cm}p{3.0cm}}
\toprule
\textbf{Year} & 
\textbf{Languages} & 
\textbf{Feature type} & 
\textbf{Feature} & 
\textbf{Reference} &
\textbf{System A} &
\textbf{System B} \\

\midrule

2020 & ru-en & POS & AUX & "This \greenemph{is} heavy oil. & "This \redemph{is} thick oil. & "It'\redemph{s} thick oil. \\

\midrule

2018 & tr-en & POS & ORDINAL & \greenemph{Thirdly}, technology is developing very fast. & \greenemph{Thirdly}, technology is evolving rapidly. & \redemph{Third}, technology is evolving rapidly. \\

\midrule

2018 & tr-en & POS & ORDINAL & The \greenemph{first} part was the repairing of the mosque, the main building. & The \greenemph{first} part was the repair of the mosque, the main building. & The \greenemph{first} part was the renovation of the main building. \\

\midrule

2018 & tr-en & POS & ADJ & \greenemph{Single} digit inflation & Inflation is \greenemph{single} digits & Inflation is the \redemph{only} household \\

\midrule

2018 & tr-en & POS & ADJ & Clearly, the murders have a \greenemph{chilling} effect. & The killings clearly had a \greenemph{chilling} effect. & The killings have clearly had a \redemph{cold} shower effect. \\

\midrule

2019 & fi-en & NER & LOC & Daytime temperatures are between 7 and 12 degrees Celsius, but cooler in \greenemph{Northern Lapland}. & Daytime temperatures are between + 7 and + 12 degrees, it's cooler in \greenemph{northern Lapland}. & Daytime temperatures are between + 7 and + 12 degrees, the North is cooler \redemph{Lapland}. \\

\midrule

2019 & fi-en & NER & LOC & It was still peaceful at least in \greenemph{Crete}, she said early on Saturday evening. & It was still peaceful, at least in \greenemph{Crete}, "he said on Saturday at the beginning of the evening. & At least there was still calm in \greenemph{Crete}, "he told the crowd in the early evening on Saturday. \\

\bottomrule
\end{tabular}

\caption{Example sentences from WMT's submissions. System A has a lower MuLER score than system B. We indicate whether the chosen feature is \greenemph{consistent} or \redemph{inconsistent} with the reference.}
\label{table_qualitative_appendix}
\end{table*}

\begin{table*}[!t]
\centering
\footnotesize
\renewcommand{\tabcolsep}{0.15cm}
\begin{tabular}{ccc}
\toprule
\textbf{POS tags} & 
\textbf{named entities} & 
\textbf{features} \\

\midrule

NOUN & TIME & GENDER \\
VERB & WORK\_OF\_ART & DEFINITE \\
PUNCT & PERSON & NUMBER \\
PROPN & NORP & \\
INTJ & CARDINAL & \\
NUM & MONEY & \\
PRON & EVENT & \\
SYM & ORDINAL & \\
SCONJ & DATE & \\
ADJ & FAC & \\
ADP & ORG & \\
ADV & LAW & \\
AUX & PRODUCT & \\
X & PERCENT & \\
CCONJ & QUANTITY & \\
DET & LANGUAGE & \\
& GPE & \\
& LOC & \\
\bottomrule
\end{tabular}
\caption{Features we use in the paper.}
\label{app:tab:feat}
\end{table*}

\begin{figure*}[tbh]  
      \centering
      \begin{subfigure}{.99\textwidth}
      \includegraphics[width=1\linewidth]{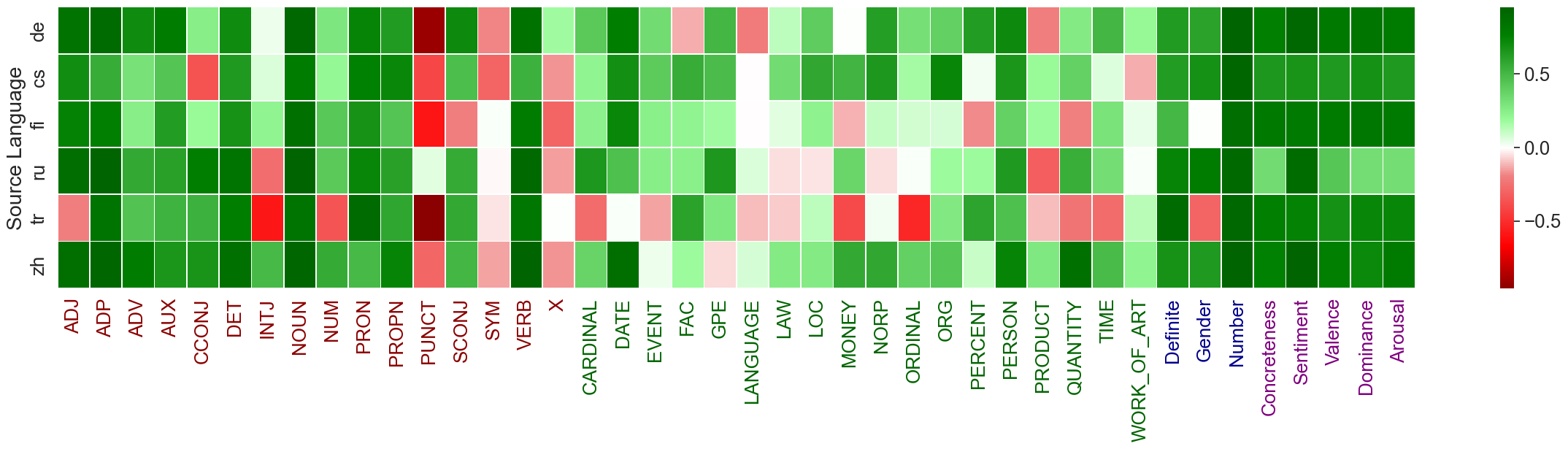}
      \caption{MuLER vs. System Sentence BLEU}
    \end{subfigure}
    \begin{subfigure}{.99\textwidth}

    \end{subfigure}
    \begin{subfigure}{.99\textwidth}
      \includegraphics[width=1\linewidth]{heat_maps/muler_VS_max_minus_min_sim.png}
      \caption{MuLER vs. Max BLEU - Min BLEU }
    \end{subfigure}
    \caption{Similarity of Measures. Represents correlation of score achievements, e.g. positive values between BLEU and MuLER suggest that BLEU increases as MuLER decreases and vice versa.}
    \label{app:fig:bleu_sim}
\end{figure*}


\begin{figure*}[tbh]
    \begin{center}
      \includegraphics[width=.99\textwidth]{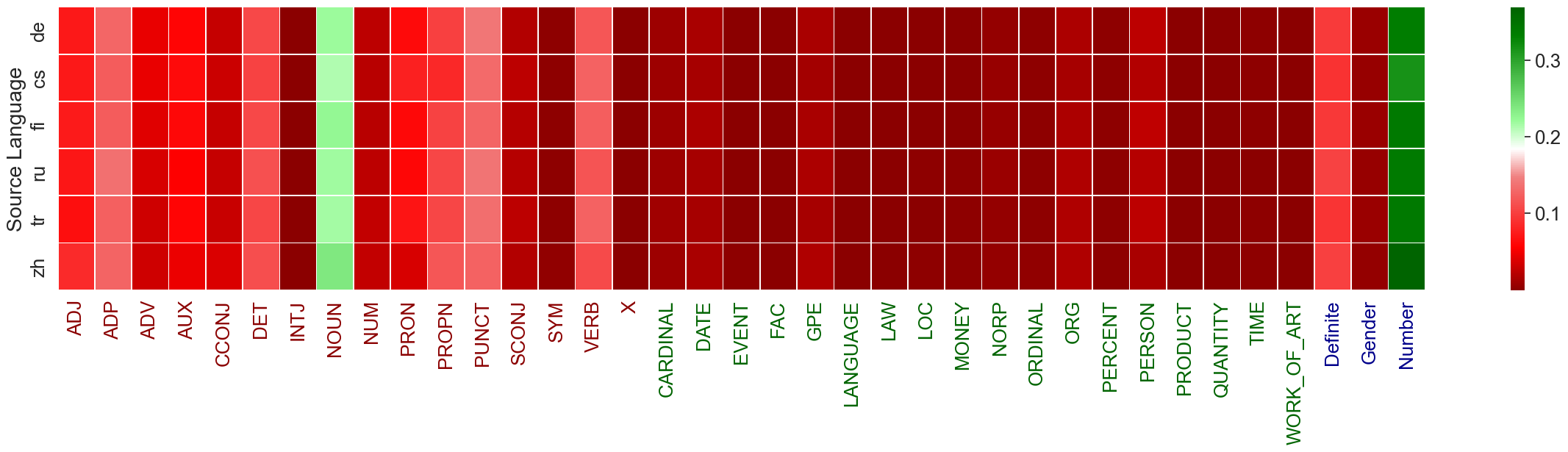}
    \end{center}
    \caption{Frequency of MuLER entities. For each language pair we chose the submission with the best BLEU score (from WMT $2014-2020$) and calculated the average frequency for each feature.}
    \label{ap:fig:feat_freq}
\end{figure*}


\begin{figure*}[tbh]
    \begin{center}
      \includegraphics[width=.99\textwidth]{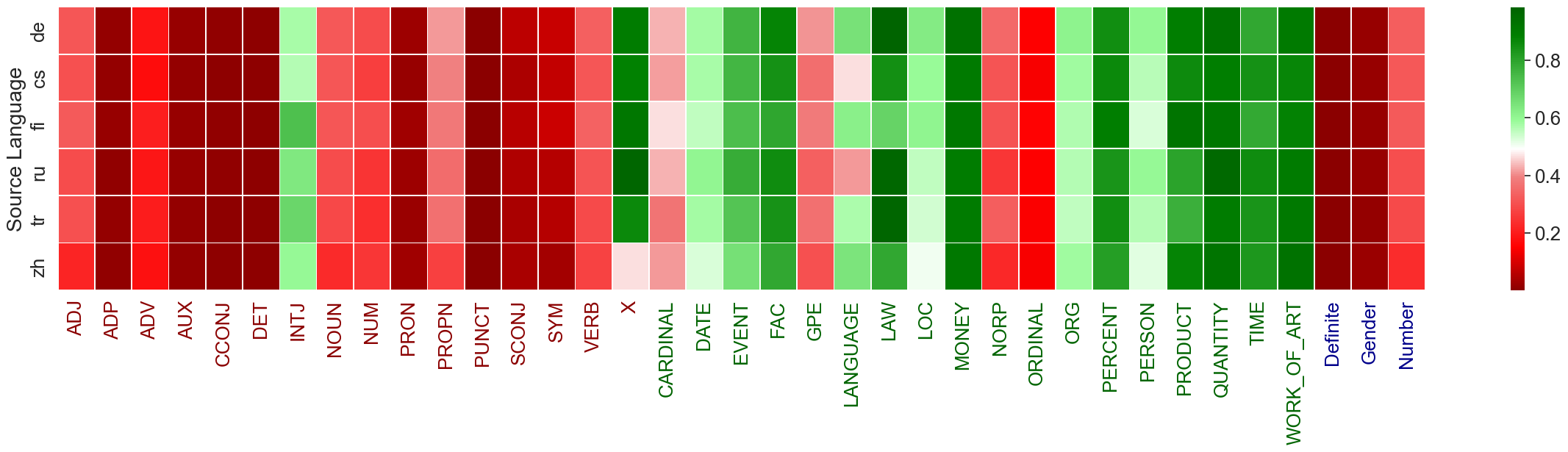}
    \end{center}
    \caption{Uniqeness of MuLER entities. For each language pair we choose the submission with the best BLEU score (from WMT $2014-2020$). For each feature we calculate its average uniqueness, defined as the number of unique times the feature appears in the text, divided by the total times it appears in the text.}
    \label{ap:fig:feat_uniq}
\end{figure*}

\begin{table*}[!t]
\centering
\footnotesize
\renewcommand{\tabcolsep}{0.15cm}
\begin{tabular}{cp{1cm}p{1.5cm}ccccccccc}
\toprule
\textbf{year} & 
\textbf{L1-L2} & 
\textbf{feature} & 
\textbf{system A} & 
\textbf{system B} & 
\textbf{A=B} & 
\textbf{A>B} & 
\textbf{B>A} & 
\thead{\textbf{MuLER} \\ \textbf{A}} & 
\thead{\textbf{MuLER} \\ \textbf{B}} & 
\thead{\textbf{BLEU} \\ \textbf{indices} \\ \textbf{A}} & 
\thead{\textbf{BLEU} \\ \textbf{indices} \\ \textbf{B}} \\

\midrule
19 & 
fi-en & 
LOC & 
\thead{newstest2019. \\ GTCOM- \\ Primary. 6946.fi-en} & 
\thead{newstest2019. \\ USYD. \\ 6995.fi-en} & 
30 & 
23 & 
1 & 
0.30 & 
0.32 & 
0.18 & 
0.32 \\

\midrule

18 & 
tr-en & 
ORDINAL & 
\thead{newstest2018. \\ online- \\ A.0.tr-en} & 
\thead{newstest2018. \\ online- \\ G.0.tr-en} & 
11 & 
13 & 
0 & 
0.06 & 
0.15 & 
0.22 & 
0.24 \\

\midrule

20 & 
ru-en & 
AUX & 
\thead{newstest2020.ru- \\ en.Online-G. \\1567} & 
\thead{newstest2020.ru- \\ en.eTranslation. \\ 686} & 
31 & 
16 & 
3 & 
0.14 & 
0.20 & 
0.34 & 
0.34 \\

\midrule

20 & 
zh-en & 
PERSON & 
\thead{newstest2020.zh- \\ en.OPPO.1422} & 
\thead{newstest2020.zh- \\ en.zlabs-nlp.1176} & 
23 & 
37 & 
2 & 
0.17 & 
0.49 & 
0.22 & 
0.19 \\

\midrule

18 & 
tr-en & 
\thead{WORK\_ \\ OF\_ \\ ART} & 
\thead{newstest2018. \\ online- \\ G.0.tr-en} & 
\thead{newstest2018. \\ online- \\ G.0.tr-en} & 
2 & 
6 & 
2 & 
0.01 & 
0.44 & 
0.25 & 
0.26 \\

\bottomrule
\end{tabular}
\caption{Manual Analysis. system A is the system with a lower MuLER score (i.e, better performance on the feature). \emph{A=B/A>B/A<B} indicates the number of sentences where the translation of the feature was of the same quality between system A and B (or better/worse accordingly). \emph{BLEU indices A/B} is the BLEU score of system A/B on sentences in the reference and the output that contain the feature.}
\label{app:tab:manual_analysis}
\end{table*}

\begin{table*}[t]
\centering
\footnotesize
\begin{tabular}{p{0.3cm}p{0.75cm}p{4cm}p{0.9cm}p{0.4cm}p{0.4cm}p{0.4cm}p{0.4cm}p{0.4cm}p{0.4cm}p{0.4cm}p{0.4cm}p{0.4cm}p{0.4cm}}
\toprule
\textbf{year} &
\textbf{langs} &
\textbf{submission} &
\textbf{system bleu} & 
\multicolumn{2}{c}{\textbf{bleu indices}} &
\multicolumn{2}{c}{\textbf{MuLER}} &
\multicolumn{2}{c}{\textbf{O}} & 
\multicolumn{2}{c}{\textbf{AO}} &
\multicolumn{2}{c}{\textbf{hybrid}} \\
\cmidrule(lr){5-6}\cmidrule(lr){7-8}\cmidrule(lr){9-10}\cmidrule(lr){11-12}\cmidrule(lr){13-14}
 & & & & noun & verb & noun & verb & noun & verb & noun & verb &
 noun & verb \\
\midrule

20 & 
de-en & 
newstest2020.de-en.OPPO.1360 & 
0.39 & 
0.41 & 
0.41 & 
0.18 & 
0.29 & 
0.45 & 
0.45 & 
0.21 & 
0.32 & 
0.33 & 
0.38 \\

\midrule

15 & 
fi-en & 
\thead[l]{newstest2015.uedin- \\ syntax.4006.fi-en} & 
0.12 & 
0.12 & 
0.13 & 
0.38 & 
0.39 & 
0.17 & 
0.16 & 
0.05 & 
0.08 & 
0.10 & 
0.12 \\

\midrule

18 & 
ru-en & 
\thead[l]{newstest2018.Alibaba. \\ 5720.ru-en} & 
0.30 & 
0.30 & 
0.30 & 
0.24 & 
0.32 & 
0.35 & 
0.34 & 
0.14 & 
0.21 & 
0.24 & 
0.27 \\

\midrule

19 & 
de-en & 
\thead[l]{newstest2019.RWTH\_ \\ Aachen\_System.6818.de-en} & 
0.33 & 
0.33 & 
0.33 & 
0.21 & 
0.28 & 
0.39 & 
0.37 & 
0.14 & 
0.24 & 
0.26 & 
0.30 \\

\midrule

20 & 
ru-en & 
\thead[l]{newstest2020.ru- \\ en.Online-G.1567} & 
0.32 & 
0.33 & 
0.33 & 
0.22 & 
0.26 & 
0.38 & 
0.36 & 
0.13 & 
0.22 & 
0.26 & 
0.28 \\

\bottomrule
\end{tabular}
\caption{Range and Monotonicity of MuLER. Presented here are MuLER scores on nouns and verbs in $5$ randomly chosen systems from WMT. Oracle (O) and Anti-Oracle (AO) masking strategies vs. hybrid masking strategy (as described in \S\ref{sec:validation}) at $50-50$ split ($50 \%$ of noun/verb is masked with O-strategy, and the rest with AO-strategy).  
\label{app:tab:hybrid_table}}
\end{table*}

\begin{table*}[t]
\centering
\footnotesize
\begin{tabular}{p{0.3cm}p{0.75cm}p{4cm}p{0.9cm}p{0.4cm}p{0.4cm}p{0.4cm}p{0.4cm}p{0.4cm}p{0.4cm}p{0.4cm}p{0.4cm}p{0.4cm}p{0.4cm}}
\toprule
\textbf{year} &
\textbf{langs} &
\textbf{submission} &
\textbf{system bleu} & 
\multicolumn{2}{c}{\textbf{bleu indices}} &
\multicolumn{2}{c}{\textbf{MuLER}} &
\multicolumn{2}{c}{\textbf{O}} & 
\multicolumn{2}{c}{\textbf{AO}} &
\multicolumn{2}{c}{\textbf{hybrid}} \\
\cmidrule(lr){5-6}\cmidrule(lr){7-8}\cmidrule(lr){9-10}\cmidrule(lr){11-12}\cmidrule(lr){13-14}
 & & & & noun & verb & noun & verb & noun & verb & noun & verb &
 noun & verb \\
\midrule

20 & 
de-en & 
newstest2020.de-en.OPPO.1360 & 
0.39 & 
0.41 & 
0.41 & 
0.18 & 
0.29 & 
0.45 & 
0.45 & 
0.21 & 
0.32 & 
0.31 & 
0.36 \\

\midrule
15 & 
fi-en & 
\thead[l]{newstest2015.uedin- \\ syntax.4006.fi-en} & 
0.12 & 
0.12 & 
0.13 & 
0.38 & 
0.39 & 
0.17 & 
0.16 & 
0.05 & 
0.08 & 
0.10 & 
0.12 \\

\midrule

18 & 
ru-en & 
\thead[l]{newstest2018.Alibaba. \\ 5720.ru-en} & 
0.30 & 
0.30 & 
0.30 & 
0.24 & 
0.32 & 
0.35 & 
0.34 & 
0.14 & 
0.21 & 
0.23 & 
0.26 \\

\midrule

19 & 
de-en & 
\thead[l]{newstest2019.RWTH\_ \\ Aachen\_System.6818.de-en} & 
0.33 & 
0.33 & 
0.33 & 
0.21 & 
0.28 & 
0.39 & 
0.37 & 
0.14 & 
0.24 & 
0.25 & 
0.30 \\

\midrule

20 & 
ru-en & 
\thead[l]{newstest2020.ru- \\ en.Online-G.1567} & 
0.32 & 
0.33 & 
0.33 & 
0.22 & 
0.26 & 
0.38 & 
0.36 & 
0.13 & 
0.22 & 
0.25 & 
0.28 \\

\bottomrule
\end{tabular}
\caption{Range and Monotonicity of MuLER. Presented here are MuLER scores on nouns and verbs in $5$ randomly chosen systems from WMT. Oracle (O) and Anti-Oracle (AO) masking strategies vs. hybrid masking strategy (as described in \S\ref{sec:validation}) at $40-60$ split ($40 \%$ of noun/verb is masked with O-strategy, and the rest with AO-strategy}

\label{app:tab:hybrid_muler4}
\end{table*}

\begin{table*}[t]
\centering
\footnotesize
\begin{tabular}{p{0.3cm}p{0.75cm}p{4cm}p{0.9cm}p{0.4cm}p{0.4cm}p{0.4cm}p{0.4cm}p{0.4cm}p{0.4cm}p{0.4cm}p{0.4cm}p{0.4cm}p{0.4cm}}
\toprule
\textbf{year} &
\textbf{langs} &
\textbf{submission} &
\textbf{system bleu} & 
\multicolumn{2}{c}{\textbf{bleu indices}} &
\multicolumn{2}{c}{\textbf{MuLER}} &
\multicolumn{2}{c}{\textbf{O}} & 
\multicolumn{2}{c}{\textbf{AO}} &
\multicolumn{2}{c}{\textbf{hybrid}} \\
\cmidrule(lr){5-6}\cmidrule(lr){7-8}\cmidrule(lr){9-10}\cmidrule(lr){11-12}\cmidrule(lr){13-14}
 & & & & noun & verb & noun & verb & noun & verb & noun & verb &
 noun & verb \\
\midrule

20 & 
de-en & 
newstest2020.de-en.OPPO.1360 & 
0.39 & 
0.41 & 
0.41 & 
0.18 & 
0.29 & 
0.45 & 
0.45 & 
0.21 & 
0.32 & 
0.31 & 
0.36 \\

\midrule

15 & 
fi-en & 
\thead[l]{newstest2015.uedin- \\ syntax.4006.fi-en} & 
0.12 & 
0.12 & 
0.13 & 
0.38 & 
0.39 & 
0.17 & 
0.16 & 
0.05 & 
0.08 & 
0.09 & 
0.11 \\

\midrule

18 & 
ru-en & 
\thead[l]{newstest2018.Alibaba. \\ 5720.ru-en} & 
0.30 & 
0.30 & 
0.30 & 
0.24 & 
0.32 & 
0.35 & 
0.34 & 
0.14 & 
0.21 & 
0.22 & 
0.26 \\

\midrule

19 & 
de-en & 
\thead[l]{newstest2019.RWTH\_ \\ Aachen\_System.6818.de-en} & 
0.33 & 
0.33 & 
0.33 & 
0.21 & 
0.28 & 
0.39 & 
0.37 & 
0.14 & 
0.24 & 
0.24 & 
0.29 \\

\midrule

20 & 
ru-en & 
\thead[l]{newstest2020.ru- \\ en.Online-G.1567} & 
0.32 & 
0.33 & 
0.33 & 
0.22 & 
0.26 & 
0.38 & 
0.36 & 
0.13 & 
0.22 & 
0.24 & 
0.28 \\

\bottomrule
\end{tabular}
\caption{Range and Monotonicity of MuLER. Presented here are MuLER scores on nouns and verbs in $5$ randomly chosen systems from WMT. Oracle (O) and Anti-Oracle (AO) masking strategies vs. hybrid masking strategy (as described in \S\ref{sec:validation}) at $30-70$ split ($30 \%$ of noun/verb is masked with O-strategy, and the rest with AO-strategy}
\label{app:tab:hybrid_muler3}
\end{table*}

\begin{table*}[!t]
\centering
\footnotesize
\renewcommand{\tabcolsep}{0.15cm}
\begin{tabular}{ccccccc|ccc}
\toprule
\multicolumn{7}{c|}{\textbf{synthetic features}} & \multicolumn{3}{c}{\textbf{features}} \\
\cmidrule(lr){1-7}\cmidrule(lr){8-10}
\thead{\textbf{average} \\ \textbf{proportion} \\ \textbf{(reference)}} & 
\thead{\textbf{average} \\ \textbf{proportion} \\ \textbf{(output)}} &
\thead{\textbf{variance of} \\ \textbf{average} \\ \textbf{proportion} \\ \textbf{(reference)}} &
\thead{\textbf{variance of} \\ \textbf{average} \\ \textbf{proportion} \\ \textbf{(output)}} &
\thead{\textbf{average} \\ \textbf{MuLER}} & 
\thead{\textbf{variance} \\ \textbf{MuLER}} & 
\thead{\textbf{std} \\ \textbf{MuLER}} & 
\textbf{feature} & 
\thead{\textbf{average} \\ \textbf{proportion}} & 
\textbf{MuLER} \\

\midrule
\greenmark{0.22} & 
\greenmark{0.22} & 
4.61e-04 &
2.57e-04 &
\redmark{0.44} & 
4.09e-04 & 
0.01 & 
NOUN & 
\greenmark{0.22} & 
\redmark{0.26} \\

\midrule

\greenmark{0.15} & 
\greenmark{0.15} &
4.86e-04 &
7.25e-04 &
\redmark{0.22} & 
2.24e-04 & 
0.01 & 
VERB & 
\greenmark{0.12} & 
\redmark{0.29} \\

\midrule

\greenmark{0.11} & 
\greenmark{0.11} & 
3.39e-04 &
2.90e-04 &
\redmark{0.21} & 
6.04e-04 & 
0.03 & 
PROPN & 
\greenmark{0.09} & 
\redmark{0.07} \\

\midrule

\greenmark{0.07} & 
\greenmark{0.07} & 
7.33e-04 &
7.12e-04 &
\redmark{0.21} & 
2.53e-04 & 
0.02 & 
PRON & 
\greenmark{0.07} & 
\redmark{0.16} \\

\midrule

\greenmark{0.05} & 
\greenmark{0.05} & 
6.71e-04 &
2.07e-04 &
\redmark{0.19} & 
6.15e-04 & 
0.02 & 
ADV & 
\greenmark{0.04} & 
\redmark{0.18} \\

\bottomrule
\end{tabular}
\caption{Specificity of MuLER. Comparison of \redmark{MuLER} for synthetic features ("average MuLER") with real features ("MuLER"). The two leftmost columns are the \greenmark{average proportion} of the synthetic features in the reference and output. The "average proportion" column indicates the average frequency of the features (e.g, NOUN/VERB) in the reference and the output (as described in \S\ref{sec:validation}). Submission is "online-G.0" for German-English from WMT $2019$.}
\label{app:tab:synthetic_features}
\end{table*}

\begin{table*}[t!] 
\centering
\footnotesize
\begin{tabular}{ccccccccc}
\toprule
\textbf{system} &
\multicolumn{2}{c}{\textbf{50\% abl-MuLER}} &
\multicolumn{2}{c}{\textbf{100\% abl-MuLER}} &
\multicolumn{2}{c}{\textbf{50\% MuLER}} &
\multicolumn{2}{c}{\textbf{100\% MuLER}} \\
\cmidrule(lr){2-3}\cmidrule(lr){4-5}\cmidrule(lr){6-7}\cmidrule(lr){8-9}
 & noun & verb & noun & verb & noun & verb & noun & verb \\
\midrule
Facebook\_FAIR.6750 &
0.021 & 0.018 & 0.054 & 0.034 & 0.203 & 0.320 & 0.267 & 0.391 \\ 
\midrule
online-A &
0.023 & 0.017 & 0.055 & 0.036 & 0.229 & 0.357 & 0.295 & 0.432 \\
\midrule
UCAM.6461. &
0.023 & 0.017 & 0.054 & 0.035 & 0.220 & 0.328 & 0.279 & 0.405 \\
\midrule
uedin.6749 &
0.022 & 0.016 & 0.056 & 0.034 & 0.242 & 0.374 & 0.306 & 0.448 \\
\midrule
online-A &
0.023 & 0.017 & 0.055 & 0.036 & 0.229 & 0.357 & 0.295 & 0.432 \\
\midrule
online-B &
0.018 & 0.016 & 0.047 & 0.032 & 0.169 & 0.286 & 0.225 & 0.359 \\
\midrule
uedin.6749 &
0.022 & 0.016 & 0.056 & 0.034 & 0.242 & 0.374 & 0.306 & 0.448 \\
\bottomrule
\end{tabular}
\caption{Robustness to Feature Frequency. Presented here are $3$ submissions from WMT $2019$, translation from German to English (see Table \ref{app:tab:frequency} for more results). We compare between MuLER and abl-MuLER (MuLER's numerator -- an ablated version of MuLER) with $50 \% / 100 \%$ of nouns/verbs masked.\label{app:tab:frequency}}

\end{table*}

\end{document}